\documentclass[10pt,twocolumn,letterpaper]{article}

\usepackage{wacv}              

\usepackage{graphicx}
\usepackage{multirow}
\usepackage{amsmath}
\usepackage{amssymb}
\usepackage{booktabs}
\usepackage{float}
\usepackage{soul}
\usepackage{tikz}
\usetikzlibrary{shapes,arrows,positioning,fit}

\usepackage[pagebackref,breaklinks,colorlinks]{hyperref}

\usepackage[capitalize]{cleveref}
\crefname{section}{Sec.}{Secs.}
\Crefname{section}{Section}{Sections}
\Crefname{table}{Table}{Tables}
\crefname{table}{Tab.}{Tabs.}

\def\wacvPaperID{4} 
\def\confName{WACV}
\def\confYear{2025}

\begin{document}

\title{DaBiT: Depth and Blur informed Transformer for Video Focal Deblurring}

\author{%
  Crispian Morris,
  Nantheera Anantrasirichai,
  Fan Zhang, and 
  David Bull \\
  School of Computer Science,
  University of Bristol,
  Bristol, UK \\\tt\small{\{crispian.morris,dave.bull,n.anantrasirichai,fan.zhang,dave.bull\}@bristol.ac.uk}}

\maketitle


\begin{abstract}

In many real-world scenarios, recorded videos suffer from accidental focus blur, and while video deblurring methods exist, most specifically target motion blur or spatial-invariant blur. This paper introduces a framework optimized for the as yet unattempted task of video focal deblurring (refocusing). The proposed method employs novel map-guided transformers, in addition to image propagation, to effectively leverage the continuous spatial variance of focal blur and restore the footage. We also introduce a flow re-focusing module designed to efficiently align relevant features between blurry and sharp domains. Additionally, we propose a novel technique for generating synthetic focal blur data, broadening the model's learning capabilities and robustness to include a wider array of content. We have made a new benchmark dataset, DAVIS-Blur, available. This dataset, a modified extension of the popular DAVIS video segmentation set, provides realistic focal blur degradations as well as the corresponding blur maps. Comprehensive experiments demonstrate the superiority of our approach. We achieve state-of-the-art results with an average PSNR performance over 1.9dB greater than comparable existing video restoration methods. Our source code and the developed databases will be made available at \url{https://github.com/crispianm/DaBiT}

\end{abstract}

\section{Introduction}
\label{sec:intro}

Videos captured in challenging circumstances, such as low light, or by less experienced shooters may exhibit blurriness, including motion blur and out-of-focus blur. Specifically, in scenarios with extremely high noise, it is difficult for a camera operator or autofocus algorithm to determine if a subject is in focus. This often results in out-of-focus footage after a denoising post-process is applied. Consequently, a focal deblurring process is necessary to recover high-frequency details, perform focus correction, and improve overall visual quality. However, this remains a challenging task, even with modern deep learning models, because of the loss of information as the blur intensifies.

While traditional methods employ deconvolution-based algorithms in order to sharpen input images \cite{anant2022artificial}, advances in deep learning techniques have enabled direct learning of a mapping from blurry to sharp. Recent contributions in this field, particularly those using convolutional neural networks (CNNs), have significantly improved upon the deconvolution-based algorithms, as demonstrated by image-based contributions such as DeBlurGAN~\cite{kupyn2019deblurgan}, NAFNet~\cite{chen2022simple}, and BIPNet~\cite{dudhane2021burst}. However, a significant challenge still exists in the form of video content; while these methods may produce visually pleasing results for individual frames, they often introduce noticeable temporal inconsistencies when applied to a video sequence. 

One explanation for this is that CNNs may struggle to capture long-range dependencies within video sequences; these are highly beneficial for handling complex image degradations like blurs while retaining high-frequency details. The transformer architecture~\cite{dosovitskiy2020image} has emerged as a promising alternative for various computer vision tasks, largely due to its self-attention mechanism. The transformer's ability to learn these global dependencies and accurately model long-range interactions makes it particularly well suited for video focal deblurring, due to the inherent temporal dependencies of blur, as demonstrated by \cite{liang2022rvrt, liang2024vrt}. While these methods have been shown to be suitable for removing camera shake and motion blur, scenarios exist where motion blur is desirable, and some undesirable focal blur is present. In such cases, focal points will vary dynamically with the content, obscuring a large amount of frames in the sequence. The spatio-temporal variance inherent in this effect makes the video focal blur problem significantly more challenging to resolve than `simple' motion blur. 

\begin{figure*}[ht]
    \centering
    \includegraphics[width=\linewidth]{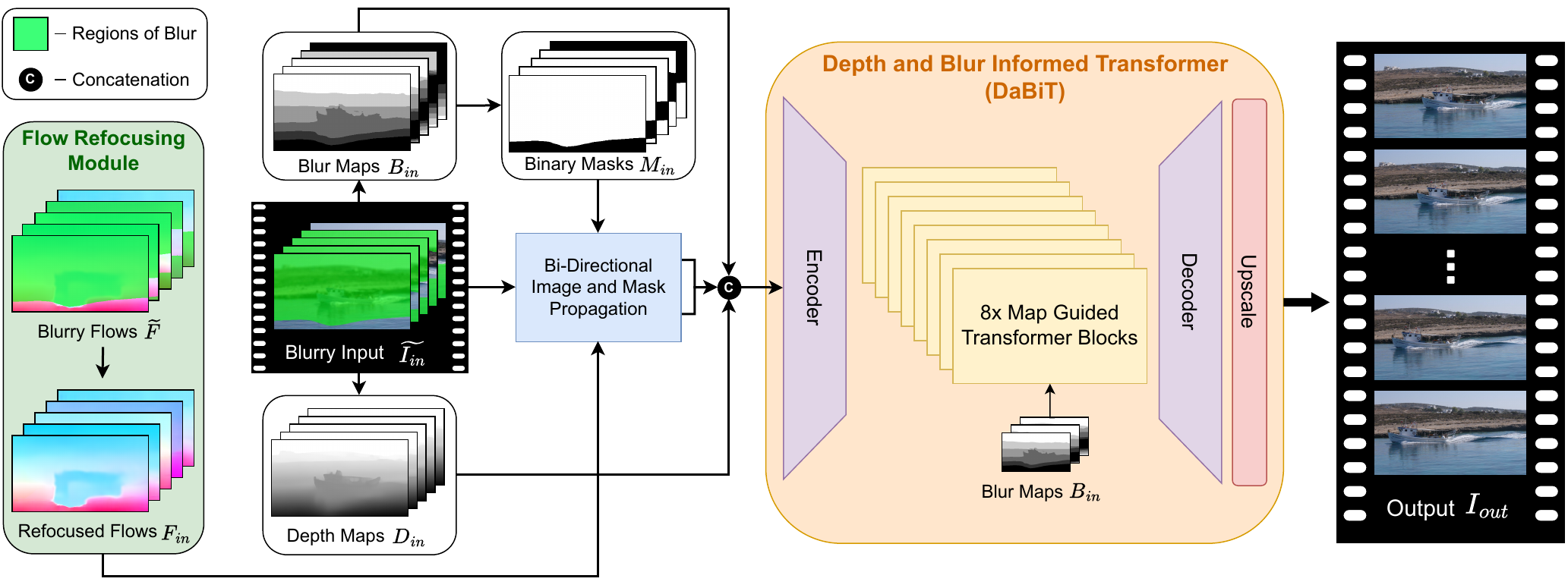}
    \caption{The architecture of the DaBiT model. Green highlights represent the temporally variant regions of the flows and input frames which are out of focus.}
    \label{fig:architecture}
\end{figure*}

In this context, this paper proposes a Depth and Blur informed Transformer (DaBiT), which aims to learn the mapping from blurry to focused. As a novel transformer-based architecture, DaBiT (illustrated in \autoref{fig:architecture}) employs depth and blur maps for video refocusing and deblurring. It also utilizes a novel spatial selection approach which reduces the computational burden, enabling recovery of information from out-of-focus areas. The DaBiT framework further incorporates a video super-resolution module, enabling flexible input resolutions. To optimize the proposed model, a large synthetic blurred video dataset is procedurally generated through randomized realistic focal blur simulation. The same generation method is also employed to create a test set, DAVIS-Blur, based on the video segmentation dataset, DAVIS \cite{perazzi2016benchmark}. The main contributions of this work are as follows:

\begin{itemize}
    \item[1)] The proposed DaBiT model is the first to attempt the task of \textbf{focus-related video deblurring}.
    
    \item[2)] We are the first to employ a \textbf{spatial selection method} to recover the areas which are out of focus, using \textbf{depth and blur maps} for added guidance.

    \item[3)] We propose a novel \textbf{focal blur simulator} and generate both training data and test dataset `\textbf{DAVIS-Blur}' for optimizing and validating focus-related video deblurring methods. Both will be released publicly along with the DaBiT model.
\end{itemize}

 The proposed DaBiT model was evaluated on our DAVIS-Blur dataset and the low-light video BVI-RLV dataset \cite{lin2024bvi}, and compared against six existing video completion, super-resolution, and refocusing models. The results demonstrated the superior performance of DaBiT, with an average PSNR gain of more than 1.9 dB.
 
\section{Related Work}
\label{sec:works}

\subsection{Video Deblurring} 

Deblurring aims to remove blur caused by incorrect focus, object motion, or camera shake from input videos. It can be accomplished either as individual images~\cite{zamir2021multi, abuolaim2020defocus, kong2023efficient, tsai2022stripformer, zamir2022restormer} or as sequences of images~\cite{hyun2015generalized, jin2018learning, zhong2023blur, kim2017dynamic, liang2024vrt, liang2022rvrt}. Importantly, only a handful of the \textit{single image} based models aim to remove blur caused by incorrect focus~\cite{abuolaim2020defocus, son2021single, ruan2022learning, zamir2022restormer, zhao2022united, cui2023image, chen2023better}. 

When methods trained to deblur individual images are applied to videos, temporal inconsistencies are prone to appear, due to small errors and inconsistencies in each image being restored differently. This necessitates the creation of specific network architectures for video deblurring. Recently, Zhong \etal proposed a Blur Interpolation Transformer~\cite{zhong2023blur} based on Swin transformer blocks, that achieved competitive performance for motion deblurring. Shang \etal attempted the joint task of frame interpolation and motion deblurring with VIDUE~\cite{shang2023joint}, improving temporal consistency significantly for the challenging 16$\times$ interpolation task. To our knowledge, no work exists aiming to remove focal blur from entire video sequences. 

\subsection{Video Super-Resolution}

Another ill-posed problem closely related to deblurring is super-resolution. Video Super-Resolution (VSR) is a well studied task, with many significant recent research contributions. It is however considered simpler than refocusing because the degradation in VSR is usually uniform across the frame. VSR is challenging when compared to image super-resolution; not only by virtue of higher computational requirements, but also due to the challenge of maintaining temporal consistency across frames. In order to remedy this, two primary approaches have been proposed: sliding window-based and recurrent-based methods. While sliding window-based VSR methods~\cite{jo2018deep, li2020mucan, tian2020tdan, li2023simple} estimate the upscaled frames using adjacent frames in the pixel space, recurrent-based methods~\cite{huang2015bidirectional, liu2022learning, chan2021basicvsr, chan2022basicvsr++} propagate latent features sequentially to synthesize the HR target frame. BasicVSR++ has emerged as a prominent tool for this task, due in large part to the effectiveness of its proposed second-order grid propagation and flow-guided deformable alignment \cite{chan2022basicvsr++}.

No specific model architecture has yet dominated, and a variety of different model backbones have been utilized for this task. In 2023, Zhou \etal proposed Upscale-A-Video~\cite{zhou2023upscale} introducing the concept of using text-guided diffusion for video upscaling, achieving a high perceptual quality. Most recently, Xu \etal proposed VideoGigaGAN~\cite{xu2024videogigagan}, balancing the inherent conflict of preserving high-frequency details whilst maintaining temporal coherence through a high-frequency shuffling mechanism. 

\begin{figure*}[!t]
    \centering
    \includegraphics[width=\linewidth]{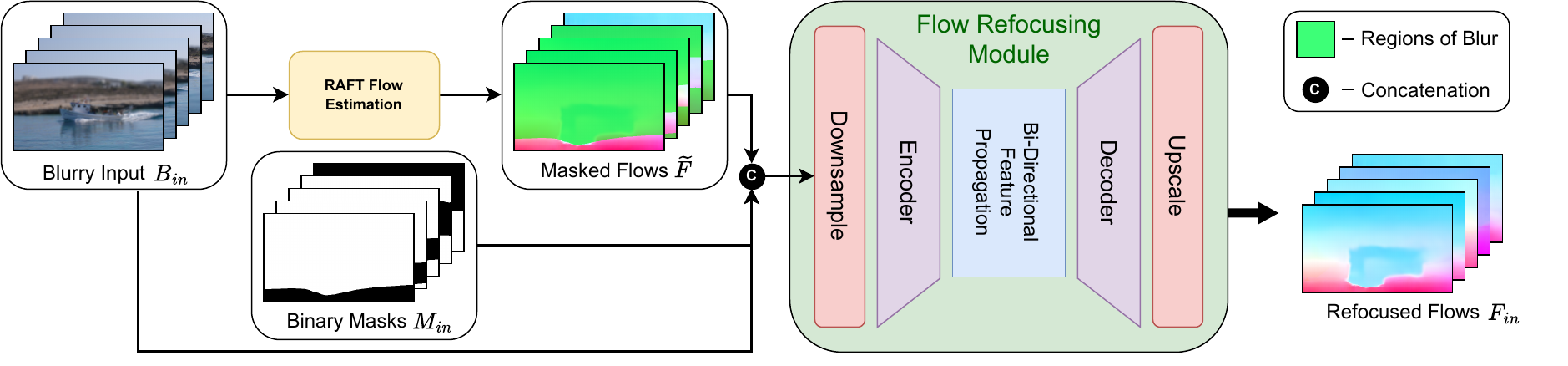}
    \caption{The architecture of the Flow Refocusing module. Green highlights represent regions of the flows which are out of focus.}
    \label{fig:flowrest}
\end{figure*}

\section{Methodology}
\label{sec:methods}

\subsection{Overview of DaBiT}
The proposed DaBiT framework is shown in \autoref{fig:architecture}. The frames and pre-computed inputs to both the model and the flow restoration module are downsampled to reduce the complexity and memory required, enabling footage of high resolution to be processed faster, also adding flexibility in other use cases, such as video editing.

Given a low-resolution, blurry input video $\widetilde{I_{in}}$ with $T$ frames and corresponding blur and depth maps $B_{in}$ and $D_{in}$, our aim is to remove the focal blur in a spatio-temporally consistent manner and jointly upscale the video to high resolution. 

The process starts with generating $B_{in}$ and the corresponding binary mask $M_{in}$ (\autoref{ssec:Blurmap}), and employs RAFT~\cite{teed2020raft} to estimate the blurry input frames' flows, $\widetilde{F} = \{\widetilde{F_t} \in \mathbb{R}^{2 \times H \times W}\}_{t=1}^{T-1}$. The flows are passed to the refocusing module (\autoref{subsec:flowrefocusing}) to obtain in-focus flows $F_{in}$. After refocusing, $\widetilde{I}_{in}$ and $M_{in}$ are passed to a non-learnable image propagation module~\cite{zhou2023propainter}, which warps flows $F_{in}$ to recover details present at different times in the sequence, using forward and backward consistency error~\cite{gao2020flow, xu2019deep}. It returns updated input frames $\widetilde{I}_{in}'$ as well as the corresponding updated masks $M_{in}'$, dilating the spatio-temporal focused area. The blur maps are left unaltered in the event that propagation delivers blurred or temporally inaccurate content.

$I_{in}', M_{in}', D_{in}$, and $B_{in}$ are then passed to the DaBiT network (\autoref{subsec:dabit}), where an encoder compresses the inputs into a latent representation, enabling a significant reduction in compute for the map guided transformers. Next, the latent features are passed into the map-guided sparse transformer blocks, processed, and decoded into frames. Finally, the deblurred frames are passed to an upscaler, which achieves 2$\times$ super-resolution, returning $I_{out}$. The addition of super-resolution at this stage gives the model the ability to add extra details lost due to blur. To minimize excess parameters, we limit super-resolution to 2$\times$, as it is experimentally found to offer an excellent trade-off between performance and complexity when compared to 4$\times$ SR.

By integrating blur and depth maps into a super-resolution transformer framework, DaBiT not only corrects out-of-focus regions but also enhances video quality, making it particularly effective for high-resolution video restoration. The specifics of each component are detailed below.

\subsection{Blur map $B_{in}$ and binary mask $M_{in}$ generation}
\label{ssec:Blurmap}
During inference, the blur maps can be estimated using a combination of the depth maps and a wavelet transform. By taking high-frequency regions of the frame found with the transform as regions of focus, we take the sum of the wavelet components for each depth, and use the inverse of this as a map of the blurred regions. These maps are needed to employ our spatial selection approach to this task, under the assumption that there are other spatio-temporally aligned regions in the video which exhibit less or no blur. This takes the form of integrating blur \textit{masks}, allowing our model to identify the regions of focus in a given input, and it reduces the difficulty of the task greatly. 

To generate the binary mask sequence $M = \{M_t \in \mathbb{R}^{1 \times H \times W}\}_{t=1}^{T}$, we begin with the blur map sequence $B_{in}$, where the intensity of the grayscale image corresponds to the amount of blur present in the spatial region. The maps are binarized by setting all but the lowest value equal to one, trivially distinguishing regions of focus from those containing blur, and delivering the resulting masks, $M_{in}$. 

\subsection{Flow Refocusing Module}
\label{subsec:flowrefocusing}

In order for the image propagation to be most effective, optical flows are required to be as close to ground truth as possible. Similarly to other models which employ pre-trained flow completion networks for inpainting objects from flows, we design a recurrent pre-trained network for aiding in the \textit{refocusing} of flows. During inference, the optical flows estimated by RAFT~\cite{teed2020raft} are between blurry, low-resolution frames. Errors in the flow have the potential to propagate into the reconstructed footage as artifacts and temporal inconsistencies. To reduce the likelihood of this, we train a separate network specifically to refocus the degraded flow, due to the distinct difference in task between refocusing flows and refocusing video sequences. We follow previous works~\cite{chan2021basicvsr, chan2022basicvsr++, zhou2023propainter} in choosing a recurrent network architecture, due to its efficiency and performance relative to a sliding window approach. 

The architecture of this module can be seen in~\autoref{fig:flowrest}. During training, we create samples in the same manner detailed in \autoref{subsec:blurmodel}, and use binarized blur maps as masks for the flows. We downsample the inputs by a factor of eight for increased speed, and pass the masked flows, binary masks, and blurry frames to an encoder. Following~\cite{chan2022basicvsr++}, we perform deformable alignment based on deformable convolution~\cite{dai2017deformable, zhu2019deformable}, propagating in-focus regions from other flows and lessening the burden of refocusing of the module. We then decode the features into refocused flows, which are used in the image propagation module to reduce the burden on the transformer.

\subsection{Map Guided Transformer}
\label{subsec:dabit}

Transformers are undoubtedly an excellent tool for video restoration, but their high memory footprint and slow inference speed pose problems in many use cases. The novel sparse transformer blocks proposed by Zhou \etal~\cite{zhou2023propainter} partially overcome this issue by building on the window-based transformer blocks proposed in E$^2$FGVI~\cite{li2022towards} and FGT~\cite{zhang2022flow}. Hence, we adopt a similar architecture to that of ProPainter, but with a few modifications. Firstly, in addition to the input frames and binary masks, we encode the depth and blur maps. This enables the transformer to learn the spatial relation between blur and depth, and ensures high similarity by encouraging consistent depth across frames. We encode this fused input into the feature space in order to further speed up processing. In contrast to ProPainter, a learnable feature propagation module is not employed before the transformer due to the minimal performance loss when ablated.

\begin{figure}[!t]
    \small
    \centering
    \includegraphics[width=\linewidth]{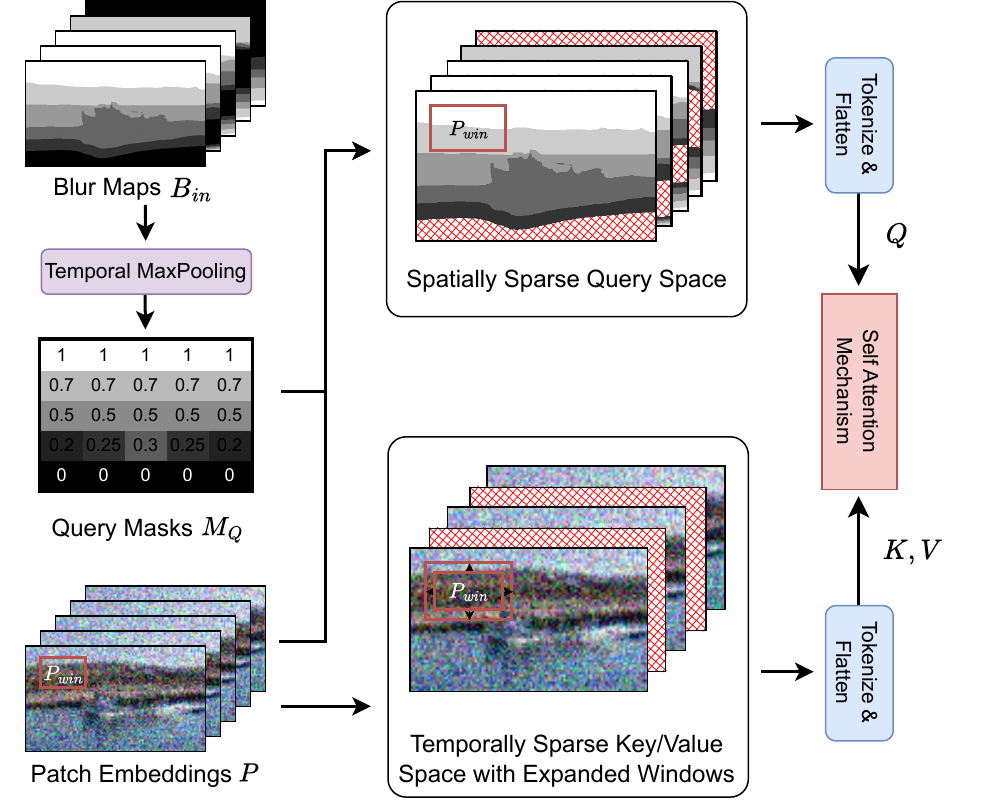}
    \caption{Map guided transformer. To decrease memory consumption and increase speed, sparsity is introduced in the key, values, and query spaces. The blur maps inform the attention mechanism of the spatio-temporal behavior of focal blur.}
    \label{fig:transformer}
\end{figure}

Soft split~\cite{liu2021fuseformer} is used to generate patch embeddings $P$ in the feature space, and decompose them into $m \times n$ non-overlapping windows of dimension $h \times w$. This results in the set of patches $P_{win}$, which we obtain the transformer's query ($Q$), key ($K$), and value ($V$) vectors from. Following~\cite{zhou2023propainter}, we design separate sparsity strategies for the query and key/value spaces as described below. We also integrate global tokens~\cite{zhang2022flow} and the window expand strategy~\cite{liu2021fuseformer} in order to replicate the $5 \times 9$ window size, the smallest possible, due to the speed achieved by ~\cite{zhou2023propainter}. An overview of this process is shown in \autoref{fig:transformer}.

\vspace{5pt}
\noindent\textbf{Query Space.} We hypothesize that communicating to the model the areas which exhibit more blur or less blur, and precisely how blurry those regions are, will aid the transformer's performance. However, compute savings are possible from any regions not exhibiting blur, since they do not need to be refocused. To this end, we downsample the blur maps and apply MaxPooling across the temporal dimension to obtain a query mask $M_{Q}$, which is zero at a pixel $(x,y)$ if the frame in question is correctly focused at that point. Crucially, the \textit{continuous} mask communicates which areas of the frames are most blurred, thereby assisting the attention mechanism in its encoding of each patch. 

\vspace{5pt}
\noindent\textbf{Key and Value Space.} The key and value space incorporates a temporal stride of two, following the same strategy as~\cite{zhou2023propainter}, effectively dividing computation evenly between each of the eight transformer blocks. This is primarily motivated by the relative similarity of blur between two given frames, and the 50\% compute savings during training and inference. 

\vspace{5pt}
\noindent\textbf{Super Resolution.} Finally, we add a simple upsampling module consisting of three cascading PixelShuffle layers to achieve 2$\times$ video super-resolution. This component allows for flexibility in input resolution without unnecessarily increasing model size or speed. 

\subsection{Loss functions}
\label{subsec:objectives}

We define three loss terms for training our model. The first is the Charbonnier loss between the model's predicted HR frames and the ground truth HR frames,
\begin{equation}
        \mathcal{L}_{Charb}(I_{pred}, I_{gt}) = \sqrt{(I_{pred}-I_{gt})^2 + \epsilon^2}, 
\end{equation}
where $I_{pred}$ and $I_{gt}$ are the predicted frames and the ground-truths, respectively, and $\epsilon = 0.001$. The second loss term applies an $l_1$ loss specifically to the blurred areas, encouraging the model to prioritize restoring these regions while leaving focused areas intact. The third term is the inverse of the second, comparing the focused input regions to the focused ground truth regions. These two loss terms are inspired by various inpainting approaches, and encourage the model to remove blur while remaining true to the in-focus regions. Thus, in total, the loss is expressed as
\begin{equation}
\begin{split}
        \mathcal{L}_{tot} &= \mathcal{L}_{Charb}(I_{pred}, I_{gt}) \\
        &+ l_1(\hat{I}_{pred}, \hat{I}_{gt}) + l_1(\tilde{I}_{pred}, \tilde{I}_{gt}),
\end{split}
\end{equation}
where $\hat{I}$ denotes only the blurred regions of the frame sequences, and $\tilde{I}$ denotes only the focused regions. Both of the two $l_1$ loss term inputs are resized to the input resolution, enabling the model to be more sensitive to the numerically small values of the Charbonnier super-resolution loss term.

\section{Training Content Generation}
\label{sec:dataset}

Deep learning necessitates a substantial volume of data to train the model effectively. While unsupervised learning techniques exist, supervised learning-based methods still outperform them in terms of both accuracy and speed. However, obtaining paired datasets of blurred and clean sequences is time-consuming and content diversity can be limited. Hence, we introduce a new approach to generate diverse synthetic training material for optimizing the proposed method. 


\subsection{Focal Blur Simulation}
\label{subsec:blurmodel}

In order for our model to learn the nature of focal blur, an accurate blur model is required. We begin with an input video $I_{gt} = \{I_t \in \mathbb{R}^{3 \times H \times W}\}_{t=1}^{T}$ with $T$ frames. We then compute the depth maps $D_{gt} = \{D_t \in \mathbb{R}^{1 \times H \times W}\}_{t=1}^{T}$ for the input using DepthAnything~\cite{depthanything}, due to its ability to predict temporally consistent depths at a level surpassing other state-of-the-art models such as MiDaS~\cite{ranftl_towards_2020} or ZoeDepth~\cite{bhat2023zoedepth}. Using the depth maps, the spatial variance of focal blur is able to be modeled. We begin by setting the focal point at a depth $f$, and defining a focal range $f_r$ centered on the focal point. We then apply gaussian blur kernels $ G(x,y) = \frac{1}{2 \pi \sigma ^2} e ^{- \frac{x^2 + y^2}{2 \sigma ^2}}$, with $\sigma = 5$, to the image in both directions along its depth axis, scaling symmetrically from zero inside the focal range to a maximum kernel size of $n_{max}\times n_{max}$. In this way, we blur the furthest depths from the focal point the most, creating a realistic focal blur effect. 

\begin{figure}[!t]
    \small
    \centering
    \includegraphics[width=\linewidth]{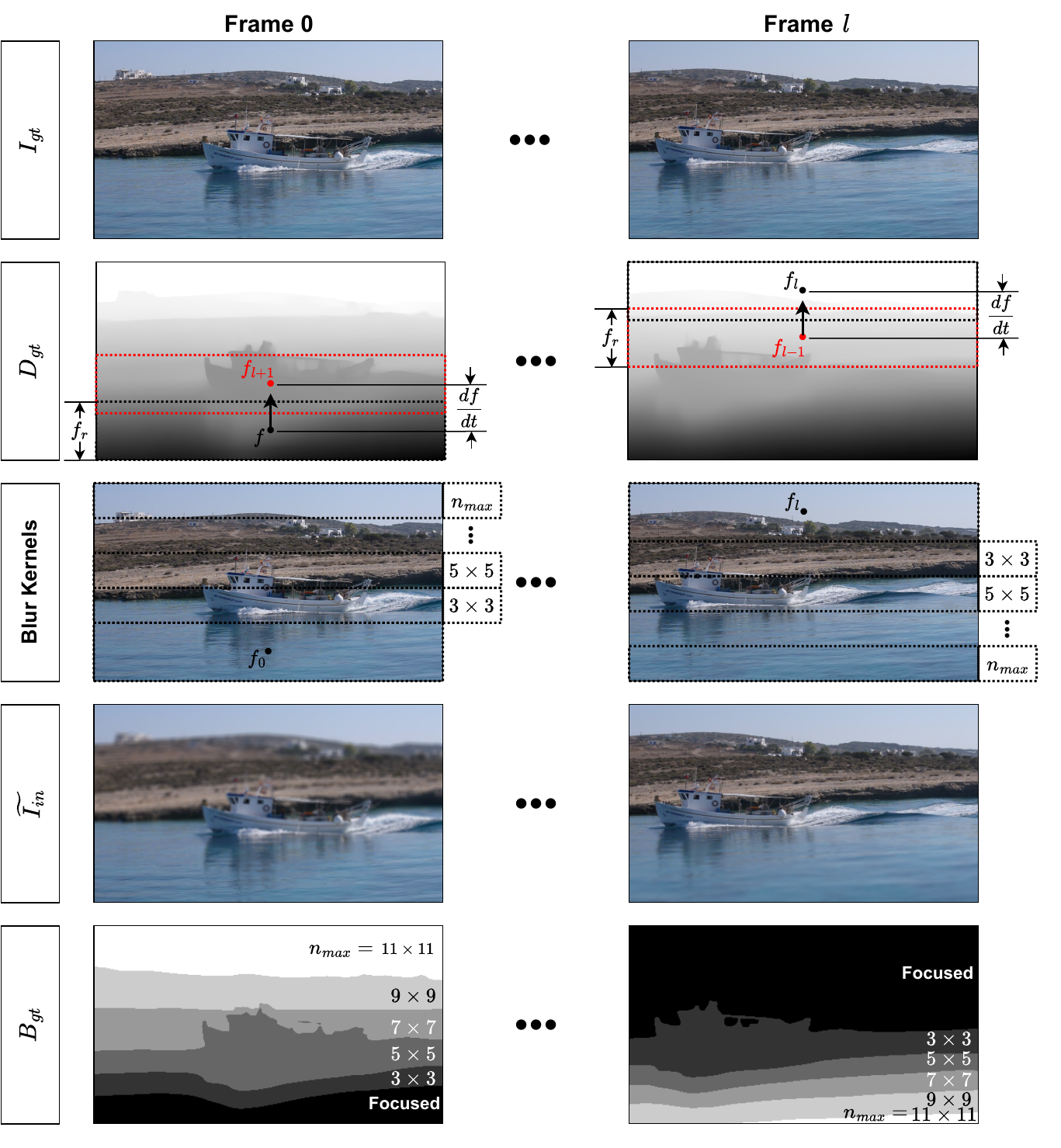}
    \caption{The training content generation process. }
    \label{fig:blurring}
\end{figure}

During training, a random frame sequence from $I_{gt}$ of length $l, l<T$, is taken, along with $r$ random global reference frames from elsewhere in the video. A random focal point ($f$) and focal range ($f_r$) are selected, along with a random focus rate $\frac{df}{dt}$, dictating the speed and direction with which the focal point will change in the sequence. A random maximum blur kernel size $n_{max} \in \{x \in \mathbb{Z} : 3 \leq x \leq 11, x \text{ is odd}\}$ is chosen, and the sequence and its global reference frames are blurred in a temporally variant manner, simulating a focus shift. The spatial blur intensity information is recorded and used to create a ground truth blur map $B_{gt} = \{B_t \in \mathbb{R}^{1 \times H \times W}\}_{t=1}^{T}$, assisting our model during training. We also introduce data augmentation by means of a random horizontal flip and random reversal of sequence during the training. Finally, to enable super-resolution, we downsample the blurred sequence, along with its blur maps, denoted as $\widetilde{I_{in}}$ and $B_{in}$, respectively. This approach is illustrated by \autoref{fig:blurring}. 

The variation in blur intensity and focus speed simulates a broad range of focus shifts. Some are representative of sequences which do not contain any additional global information (\eg $\frac{df}{dt} = 0$, so $l$ is out-of-focus in the same spatio-temporal regions), and some which contain enough focused footage to be able to nearly reconstruct the whole sequence using only image propagation (\eg $f$ is varied completely across $\text{min}(D_{gt})$ and $\text{max}(D_{gt})$). This allows a given model to learn to deblur fully out-of-focus frames while not discarding the information contained in focused frames.

\begin{table*}[ht]
    \centering
    \resizebox{1\linewidth}{!}{\begin{tabular}{r|r|ccc|cc}
        \toprule
        \multicolumn{1}{c|}{\textbf{Approach}} & \multicolumn{1}{c|}{\textbf{Model}} & \textbf{PSNR (dB) $\uparrow$} & \textbf{SSIM $\uparrow$} & \textbf{tOF $\downarrow$} & \textbf{\#P (M)} & \textbf{RT (s)} \\
        
         \midrule
        \multirow{1}{*}{Video Completion}             & ProPainter~\cite{zhou2023propainter}  & 17.962 & 0.597 & 4.992 & 39.4 & 0.107 \\
        \midrule
        \multirow{2}{*}{Video SR}               & BasicVSR++ (120p)~\cite{chan2022basicvsr++} & 25.186 & 0.679 & 3.983 & 7.3 & 0.022 \\
                                                & BasicVSR++~\cite{chan2022basicvsr++} & 25.618 & 0.713 & 3.716 & 7.3 & 0.150 \\
        \midrule
        \multirow{2}{*}{Image Refocusing}       & Restormer~\cite{zamir2022restormer}  & 24.715 & 0.698 & 2.902 & 26.5 & 0.093 \\
                                                & APL~\cite{zhao2022united} & 23.870 & 0.605 & 4.768 & 13.5 & 0.017 \\
        \midrule
        \multirow{6}{*}{Joint Refocusing}       & FMA-Net*~\cite{youk2024fma}  & 25.691 & \textcolor{blue}{\underline{0.721}} & 2.396 & 9.6 & 0.543 \\ 
                                                & FMA-Net (120p)*~\cite{youk2024fma}  & \textcolor{blue}{\underline{26.860}} & 0.720 & \textcolor{blue}{\underline{2.291}} & 9.6 & 0.151 \\ 
                                                & \textbf{DaBiT (Ours)} & \textbf{\textcolor{red}{28.777}} & \textbf{\textcolor{red}{0.811}} & \textbf{\textcolor{red}{1.239}} & 45.5 & 0.112 \\
             \cmidrule{2-7}                                  
        &V1 w/o Blur Maps                       & 25.311 & 0.706 & 2.499 &--&--\\ 
        &V2 w/o Image Propagation               & 27.203 & 0.758 & 1.914 &--&--\\
        &V3 w/o Blur Maps and Image Propagation & 25.227 & 0.703 & 2.548 &--&-- \\
        \bottomrule                                               
    \end{tabular}}
    \caption{Quantitative comparison results for DaBiT (and its ablation study variants) and three other tested methods. For each column, the best result is bold in \textbf{\textcolor{red}{red}} and the second best is underlined in \textcolor{blue}{\underline{blue}}. The average runtime in seconds (RT) for refocusing a 240p frame as well as the number of model parameters in millions (\#P) for each method are also reported. For super-resolution models, the output is downsampled to match the 480p ground truth DAVIS scene, and for non-super-resolution models the ground truth is downsampled. Superscript (*) indicates models have been retrained using our defocal blur training approach.}
    \label{tbl:quant}
\end{table*}

To generate the training material, we combined the 3471 sequences from the Youtube-VOS~\cite{xu2018youtube} and the 200 sequences of the BVI-DVC~\cite{ma2020bvi} datasets. Youtube-VOS is a large video database designed for object segmentation, while BVI-DVC contains diverse content with a wide range of low-level features and semantic scenarios, which has been employed to optimize learning-based models for various low-level tasks \cite{danier2022st,ma2020mfrnet,zhang2021video}. This results in a total of 3671 unique videos. 

\section{Experimental setup}
\label{sec:experiment}

\noindent\textbf{Training Details and Metrics.}  During the training of the Flow Refocusing module, 20 iterations of RAFT~\cite{teed2020raft} are used for flow estimation with a frame sequence length ($l$) of 10. We follow~\cite{zhou2023propainter} and apply propagation to 8$\times$ downsampled feature vectors. In the main model, eight sparse transformer blocks with four heads are employed for deblurring a local video sequence $l = 10$, with $r=6$ reference frames. The map guided transformer window size is 5$\times$9, and the extended size is half of the window size. 

Both DaBiT and the Flow Refocusing module are trained using the Adam~\cite{kingma2014adam} optimizer, selected for its robust performance in optimizing deep learning models. A fixed learning rate of $10^{-4}$ and a batch size of 1 are used for both models to ensure gradual convergence. Due to memory constraints, we train the Flow Refocusing module and DaBiT on frames of size 432$\times$240, and continue for 100k iterations and 300k iterations, respectively
\vspace{5pt}
\noindent\textbf{Benchmark methods.} We benchmark DaBiT against a variety of state-of-the-art image and video restoration models, including video inpainting and completion model  ProPainter~\cite{zhou2023propainter}, video super resolution approach BasicVSR++~\cite{chan2022basicvsr++}, two image refocusing methods,  Restormer~\cite{zamir2022restormer} and APL~\cite{zhao2022united}, and one joint refocusing model, FMA-Net~\cite{youk2024fma}. 

\vspace{5pt}

\noindent\textbf{Test datasets and metrics.} In order to create suitable test content, we apply the same content generation workflow to 50 sequences in the DAVIS~\cite{perazzi2016benchmark} database, which results in a new test set for refocusing, DAVIS-Blur. Instead of the randomized parameters used during training, fixed parameters are used to standardize the test set. We set $l = T$, $n_{max} = 11$, $f_r = 100$, fix the initial focal point $f = 0$, and configure the focus rate to be $\text{max}(D) / l$, according to the length of each sequence. This simulates a complete focus shift from near to far, across the full length of the sequence, in all 50 sequences. It is noted that the large range of sequence lengths in the DAVIS set allows a wide range of focus shifts to be synthesized, thus posing a challenging task. Additionally, DaBiT and benchmark methods were tested on real low-light scenes, which often face focusing problems due to high noise and low contrast. The BVI-RLV \cite{lin2024bvi} dataset was used to evaluate their performance in low-light scenarios.

We use PSNR and SSIM~\cite{wang2004image} to evaluate our methods, as well as tOF~\cite{chu2020learning} for measuring temporal consistency. The number of parameters in the model, expressed in millions, is reported along with the average runtime for producing a single 240p frame in seconds. The complexity figures are calculated on a PC with an Intel Core i7-13700 CPU, 64GB RAM, and a single NVIDIA RTX 4090 GPU.

\begin{figure*}[!t]
    \small
    \centering
    \begin{minipage}[b]{0.162\linewidth}
        \centering
        \includegraphics[width=\linewidth]{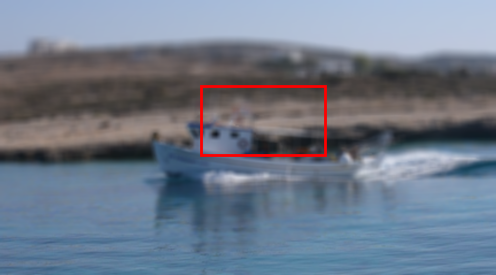}
    \end{minipage}
    \begin{minipage}[b]{0.162\linewidth}
        \centering
        \includegraphics[width=\linewidth]{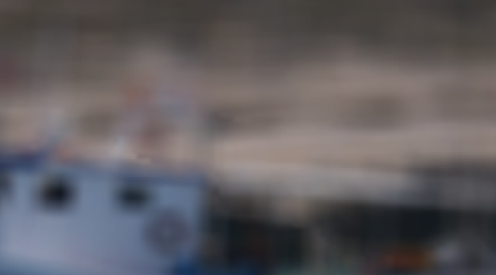}
    \end{minipage}
    \begin{minipage}[b]{0.162\linewidth}
        \centering
        \includegraphics[width=\linewidth]{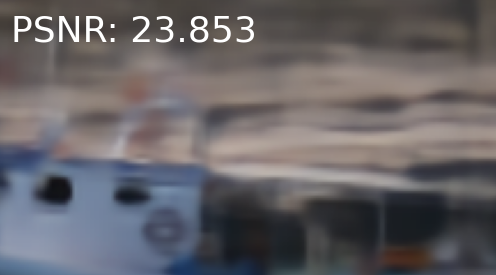}
    \end{minipage}
    \begin{minipage}[b]{0.162\linewidth}
        \centering
        \includegraphics[width=\linewidth]{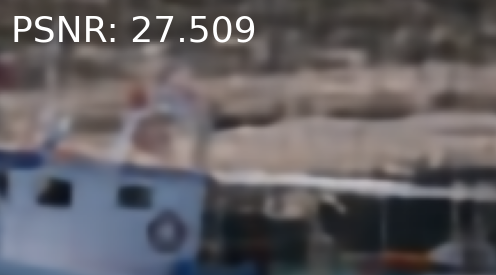}
    \end{minipage}
    \begin{minipage}[b]{0.162\linewidth}
        \centering
        \includegraphics[width=\linewidth]{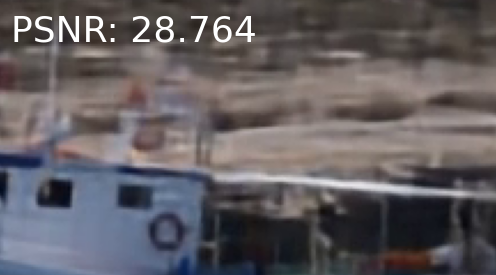}
    \end{minipage}
    \begin{minipage}[b]{0.162\linewidth}
        \centering
        \includegraphics[width=\linewidth]{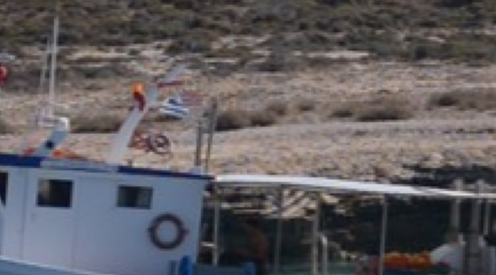}
    \end{minipage}
    \\
    \begin{minipage}[b]{0.162\linewidth}
        \centering
        \includegraphics[width=\linewidth]{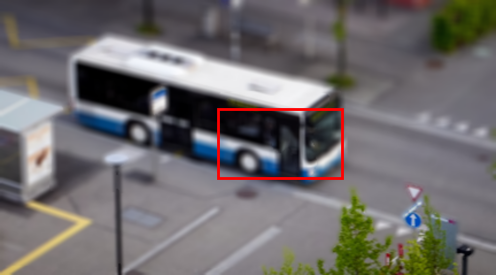}
    \end{minipage}
    \begin{minipage}[b]{0.162\linewidth}
        \centering
        \includegraphics[width=\linewidth]{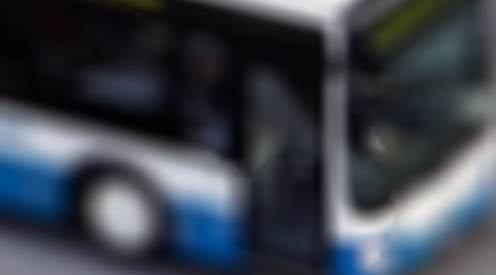}
    \end{minipage}
    \begin{minipage}[b]{0.162\linewidth}
        \centering
        \includegraphics[width=\linewidth]{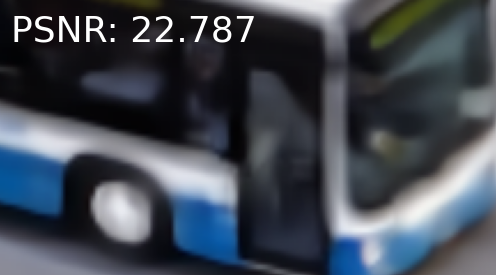}
    \end{minipage}
    \begin{minipage}[b]{0.162\linewidth}
        \centering
        \includegraphics[width=\linewidth]{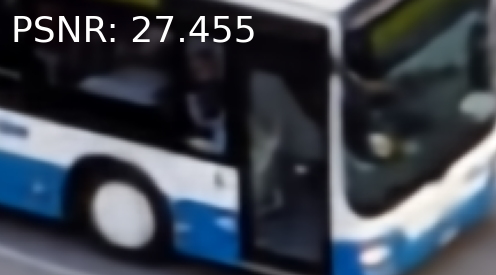}
    \end{minipage}
    \begin{minipage}[b]{0.162\linewidth}
        \centering
        \includegraphics[width=\linewidth]{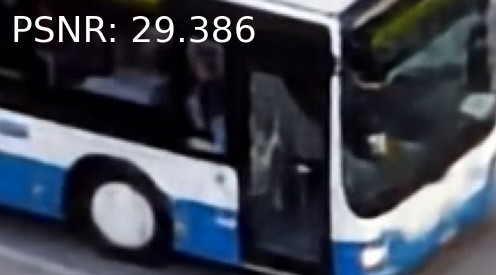}
    \end{minipage}
    \begin{minipage}[b]{0.162\linewidth}
        \centering
        \includegraphics[width=\linewidth]{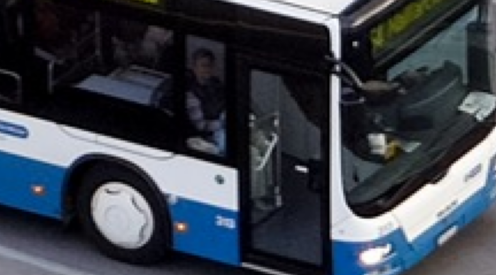}
    \end{minipage}
    \\
    \begin{minipage}[b]{0.162\linewidth}
        \centering
        \includegraphics[width=\linewidth]{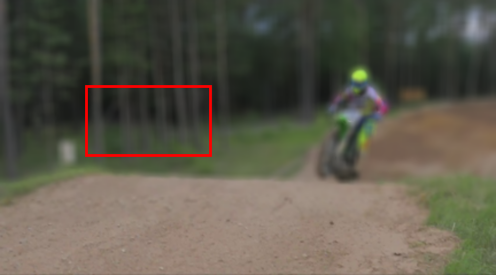}
    \end{minipage}
    \begin{minipage}[b]{0.162\linewidth}
        \centering
        \includegraphics[width=\linewidth]{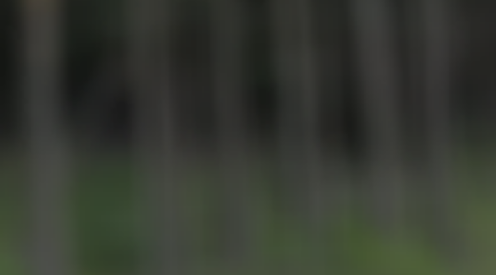}
    \end{minipage}
    \begin{minipage}[b]{0.162\linewidth}
        \centering
        \includegraphics[width=\linewidth]{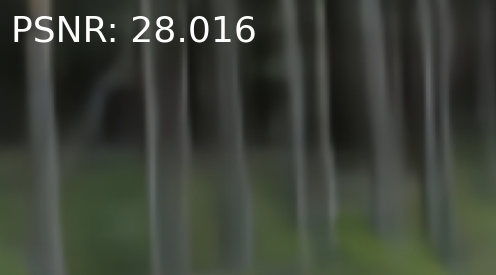}
    \end{minipage}
    \begin{minipage}[b]{0.162\linewidth}
        \centering
        \includegraphics[width=\linewidth]{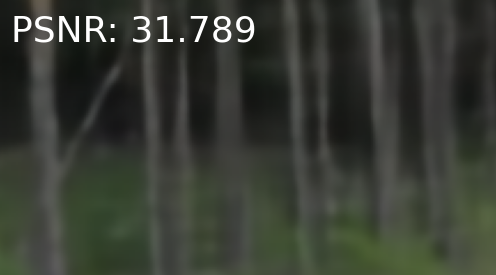}
    \end{minipage}
    \begin{minipage}[b]{0.162\linewidth}
        \centering
        \includegraphics[width=\linewidth]{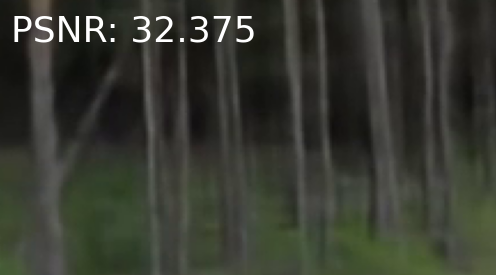}
    \end{minipage}
    \begin{minipage}[b]{0.162\linewidth}
        \centering
        \includegraphics[width=\linewidth]{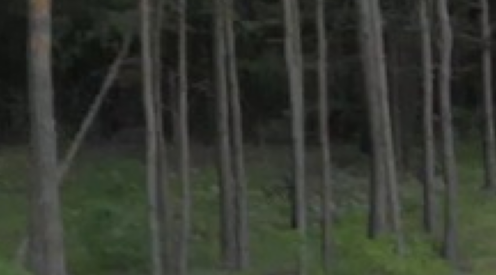}
    \end{minipage}
    \\
    \begin{minipage}[b]{0.162\linewidth}
        \centering
        \includegraphics[width=\linewidth]{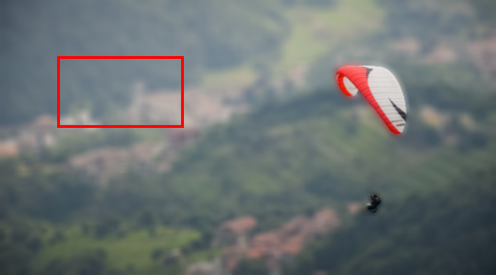}
        {(a) Input}
    \end{minipage}
    \begin{minipage}[b]{0.162\linewidth}
        \includegraphics[width=\linewidth]{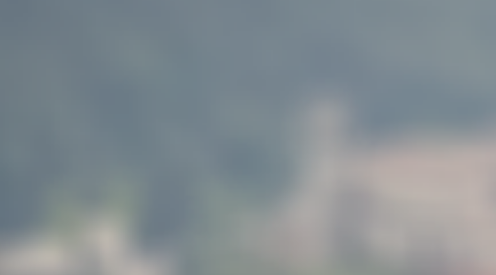}
        {(b) Input Zoom}
    \end{minipage}
    \begin{minipage}[b]{0.162\linewidth}
        \centering
        \includegraphics[width=\linewidth]{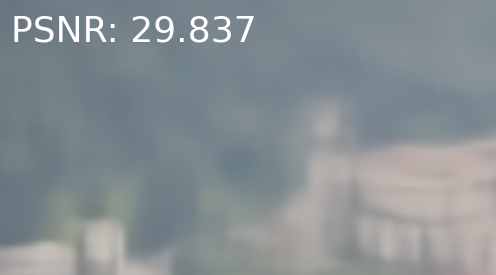}
        {(c) Restormer}
    \end{minipage}
    \begin{minipage}[b]{0.162\linewidth}
        \centering
        \includegraphics[width=\linewidth]{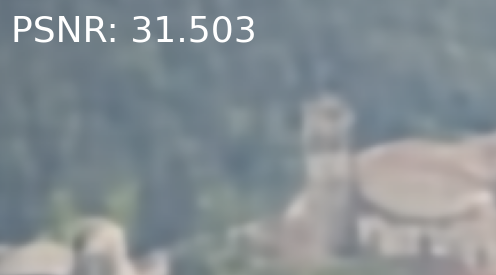}
        {(d) FMA-Net*}
    \end{minipage}
    \begin{minipage}[b]{0.162\linewidth}
        \centering
        \includegraphics[width=\linewidth]{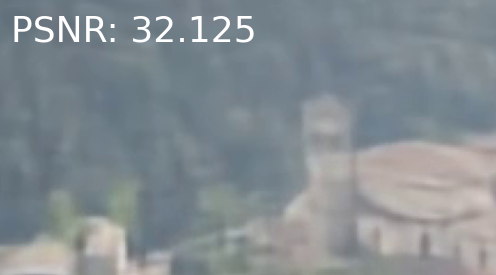}
        {(e) Ours}
    \end{minipage}
    \begin{minipage}[b]{0.162\linewidth}
        \centering
        \includegraphics[width=\linewidth]{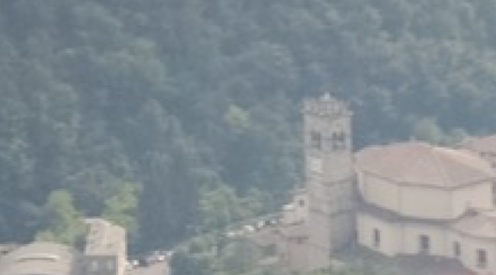}
        {(f) GT}
    \end{minipage}
    \caption{Qualitative examples from the DAVIS-Blur test set demonstrating the fine details preserved by our approach. Superscript (*) indicates models that have been retrained to remove focal blur using our training approach. }
	\label{fig:qualitative}
\end{figure*}

\begin{figure*}[!t]
    \small
    \centering
    \begin{minipage}[b]{0.162\linewidth}
        \centering
        \includegraphics[width=\linewidth]{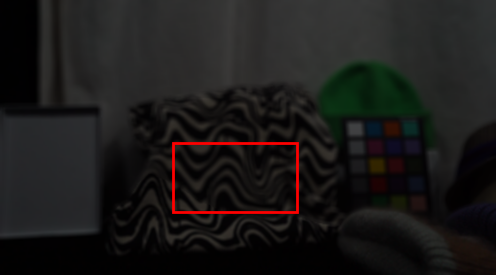}
    \end{minipage}
    \begin{minipage}[b]{0.162\linewidth}
        \centering
        \includegraphics[width=\linewidth]{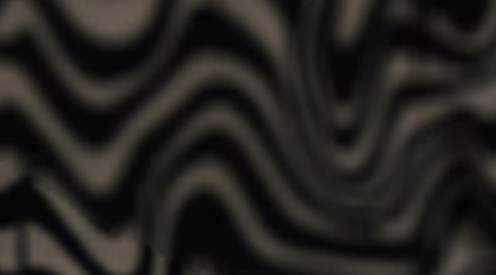}
    \end{minipage}
    \begin{minipage}[b]{0.162\linewidth}
        \centering
        \includegraphics[width=\linewidth]{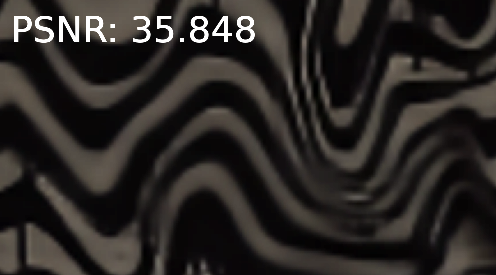}
    \end{minipage}
    \begin{minipage}[b]{0.162\linewidth}
        \centering
        \includegraphics[width=\linewidth]{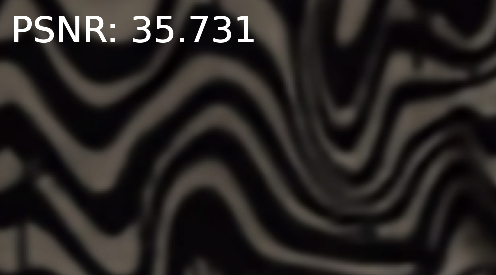}
    \end{minipage}
    \begin{minipage}[b]{0.162\linewidth}
        \centering
        \includegraphics[width=\linewidth]{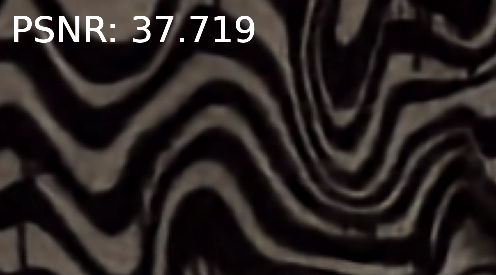}
    \end{minipage}
    \begin{minipage}[b]{0.162\linewidth}
        \centering
        \includegraphics[width=\linewidth]{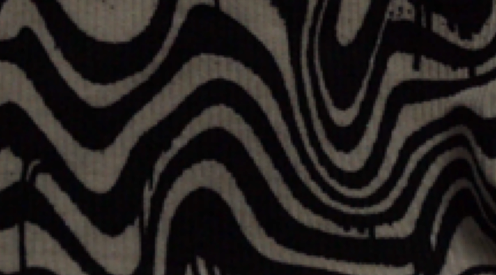}
    \end{minipage}
    \\
    \begin{minipage}[b]{0.162\linewidth}
        \centering
        \includegraphics[width=\linewidth]{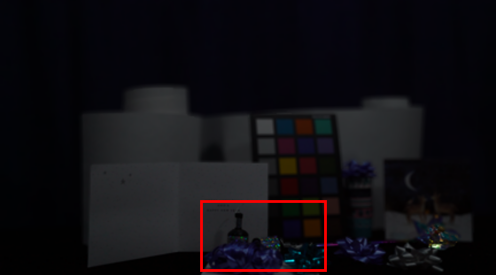}
    \end{minipage}
    \begin{minipage}[b]{0.162\linewidth}
        \centering
        \includegraphics[width=\linewidth]{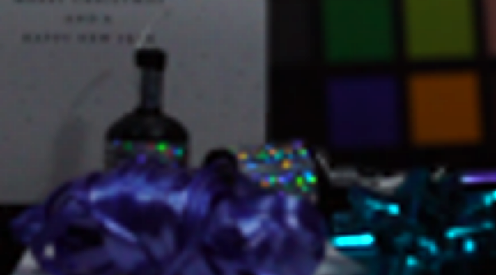}
    \end{minipage}
    \begin{minipage}[b]{0.162\linewidth}
        \centering
        \includegraphics[width=\linewidth]{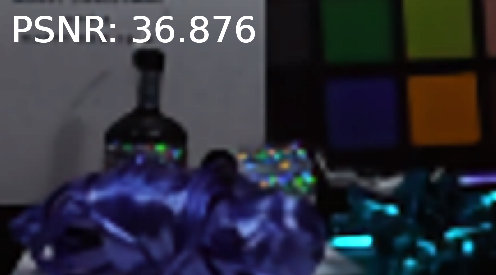}
    \end{minipage}
    \begin{minipage}[b]{0.162\linewidth}
        \centering
        \includegraphics[width=\linewidth]{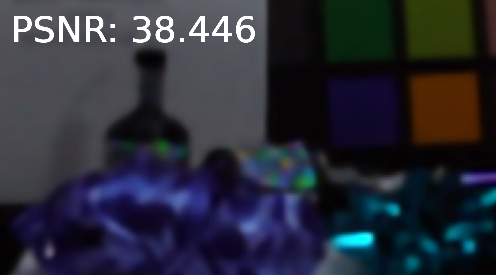}
    \end{minipage}
    \begin{minipage}[b]{0.162\linewidth}
        \centering
        \includegraphics[width=\linewidth]{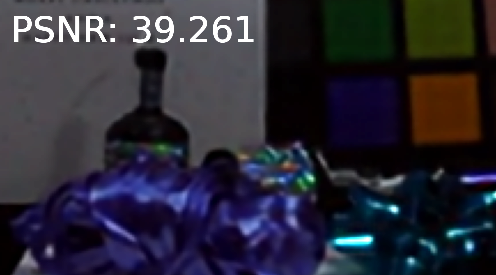}
    \end{minipage}
    \begin{minipage}[b]{0.162\linewidth}
        \centering
        \includegraphics[width=\linewidth]{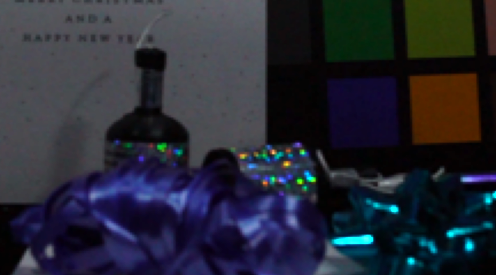}
    \end{minipage}
    \\
    \begin{minipage}[b]{0.162\linewidth}
        \centering
        \includegraphics[width=\linewidth]{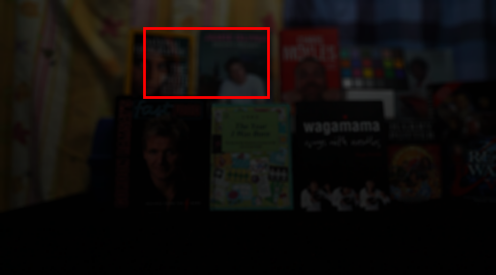}
    \end{minipage}
    \begin{minipage}[b]{0.162\linewidth}
        \centering
        \includegraphics[width=\linewidth]{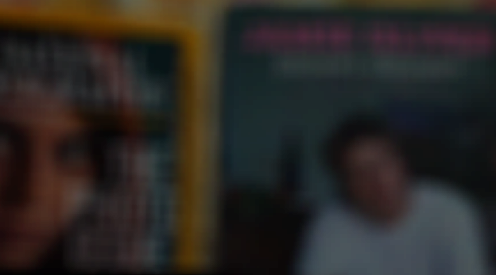}
    \end{minipage}
    \begin{minipage}[b]{0.162\linewidth}
        \centering
        \includegraphics[width=\linewidth]{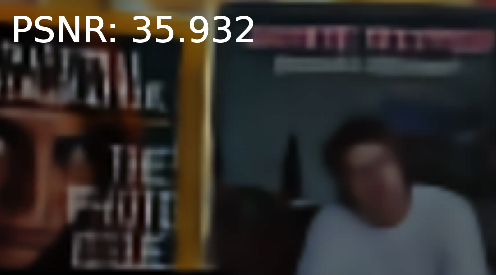}
    \end{minipage}
    \begin{minipage}[b]{0.162\linewidth}
        \centering
        \includegraphics[width=\linewidth]{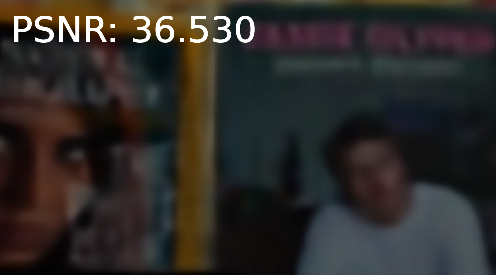}
    \end{minipage}
    \begin{minipage}[b]{0.162\linewidth}
        \centering
        \includegraphics[width=\linewidth]{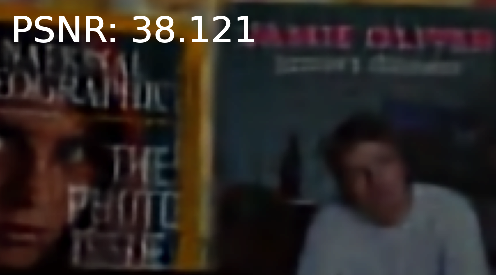}
    \end{minipage}
    \begin{minipage}[b]{0.162\linewidth}
        \centering
        \includegraphics[width=\linewidth]{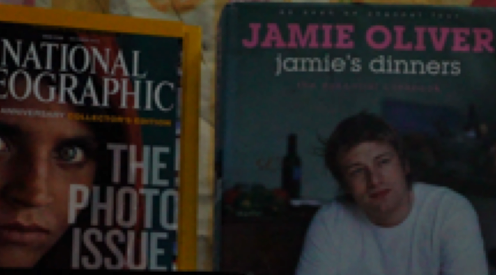}
    \end{minipage}
    \\
    \begin{minipage}[b]{0.162\linewidth}
        \centering
        \includegraphics[width=\linewidth]{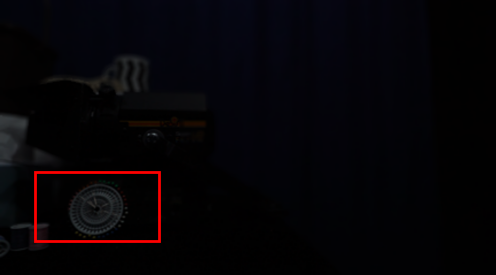}
        {(a) Input}
    \end{minipage}
    \begin{minipage}[b]{0.162\linewidth}
        \centering
        \includegraphics[width=\linewidth]{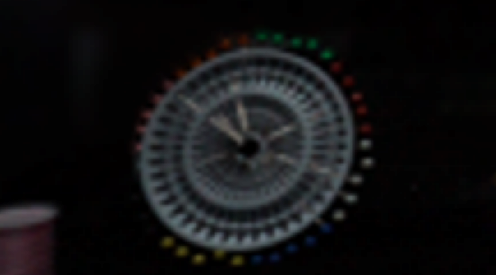}
        {(b) Input Zoom}
    \end{minipage}
    \begin{minipage}[b]{0.162\linewidth}
        \centering
        \includegraphics[width=\linewidth]{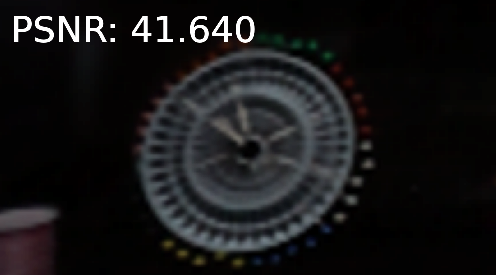}
        {(c) Restormer}
    \end{minipage}
    \begin{minipage}[b]{0.162\linewidth}
        \centering
        \includegraphics[width=\linewidth]{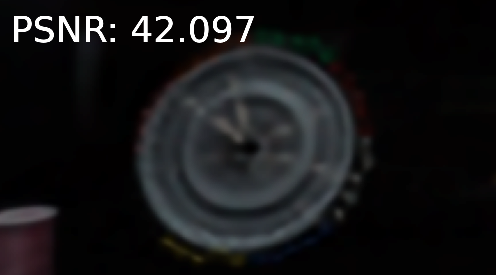}
        {(d) FMA-Net*}
    \end{minipage}
    \begin{minipage}[b]{0.162\linewidth}
        \centering
        \includegraphics[width=\linewidth]{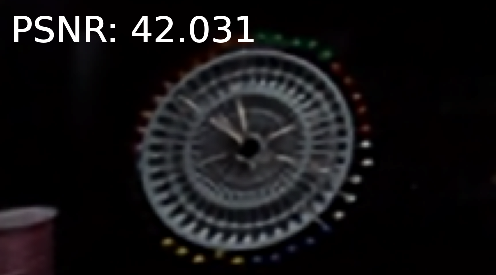}
        {(e) Ours}
    \end{minipage}
    \begin{minipage}[b]{0.162\linewidth}
        \centering
        \includegraphics[width=\linewidth]{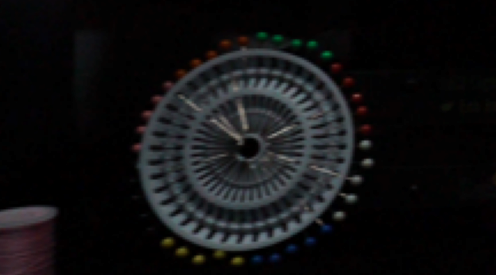}
        {(f) GT}
    \end{minipage}
    \caption{Qualitative examples from blurred BVI-RLV scenes showing the performance of our approach on lowlight data. The brightness has been increased by a factor of two for columns (b)\textendash(f) for increased visibility. Superscript (*) indicates models that have been retrained to remove focal blur using our training approach.}
	\label{fig:lowlight}
    \vspace{-10pt}
\end{figure*}

\section{Results and Discussion}

\subsection{Quantitative results} 

\autoref{tbl:quant} summarizes the comparison results between our method and other five state-of-the-art methods on the DAVIS-Blur dataset. For a fair comparison, deblurring methods not originally designed for removing defocal blur are finetuned on our data with our training approach for 35 epochs, while FMA-Net is fully retrained. Due to the 4$\times$ upscaling nature of FMA-Net and BasicVSR++, we run the models in two ways. In the first, we pass a 240p frame and downsample the output to match the ground truth, and in the second we pass a 120p frame for comparison to the 480p ground truth. Also for fairness, the APL~\cite{zhao2022united} codebase is modified to include a significant speedup, and ProPainter is run in video completion mode.

DaBiT outperforms all other tested methods by a minimum of 1.9 dB PSNR and 0.09 SSIM. In addition, it shows very high temporal consistency across frames, with a minimum decrease of 1.05 tOF from the next best model. For our method, the time taken for estimating depth is not taken into account, but the RAFT estimate time is included, making the model competitively fast, given its size.

\subsection{Qualitative results}

Visual samples of our model's outputs on DAVIS-Blur are shown in~\autoref{fig:qualitative}. It can be observed that the DaBiT outputs are considerably more detailed than those from other methods. Additionally, our model shows promise on low-light focal blur, as tested on select scenes from the BVI-RLV dataset~\cite{lin2024bvi}, shown in~\autoref{fig:lowlight}. In this case, we create a similar test environment to that of DAVIS-Blur, with an initial focal point $f = 0$, a maximum blur kernel of $7 \times 7$, and a focal range of 100.

\subsection{Ablation study}

\noindent\textbf{V1 w/o Blur Maps.} One of the novelties in this work is the use of blur maps in the proposed DaBiT model. To test its importance, we have created a variant (V1) by replacing the variable with a tensor of all ones. This is essentially telling the model that the entirety of every frame is blurred, preventing the use of learned performance from blur maps.

\vspace{5pt}

\noindent\textbf{V2 w/o Image Propagation.} To verify the effectiveness of the image propagation component, we have removed the image and mask propagation present in the model, obtaining V2. Since the image propagation module is not learned, and simply updates the masks and blurry inputs, this version of our model does not need to be retrained. 

\vspace{5pt}

\noindent\textbf{V3 w/o Blur Maps and Image Propagation.} Another variant (V3) was created by removing both blur maps and image propagation. The ablation study results are also presented in \autoref{tbl:quant}, where all three variants show inferior PSNR, SSIM and tOF performance compared to the original DaBiT model. This demonstrates the contribution of these components. Moreover, it has been observed that without blur maps (V1), the model results in a more significant performance loss (more than 3.4dB in PSNR) compared to that without image propagation (with a loss of 1.5dB in PSNR).

\subsection{Limitations}

While the proposed DaBiT model demonstrates impressive performance in video refocusing tasks, there are still several limitations that need to be addressed.

\begin{figure}[!t]
    \centering
    \includegraphics[width=\columnwidth, height=2.5cm]{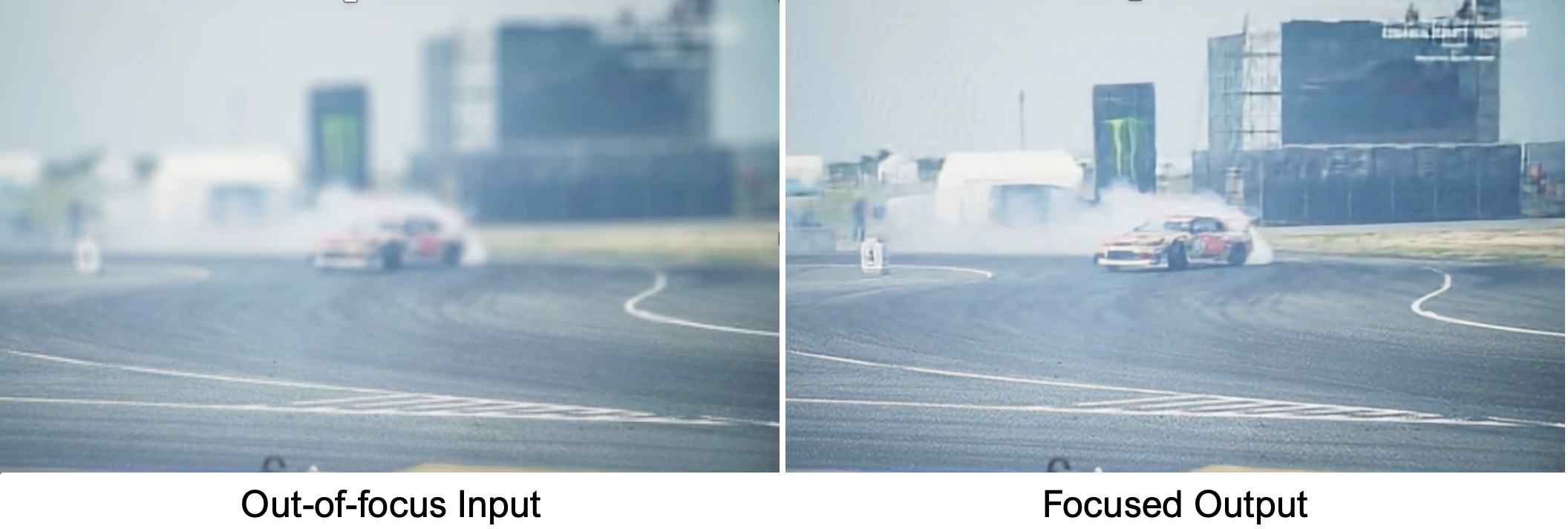}
    \caption{Refocusing result of `Car' sequence, in which the sky contains visible artifacts while other areas are enhanced with more spatial details.}
    \label{fig:limitation_car}
    \vspace{-10pt}
\end{figure}

\vspace{5pt}
\noindent\textbf{Blur map estimation.} We found that some results contain visible artifacts in homogeneous areas. This may be due to bandlimiting, which leads to inaccurate blur map estimation. An example is shown in \autoref{fig:limitation_car}. While other areas are sharpened with significantly enhanced details by our refocusing method, the sky exhibits some artifacts. However, this can be easily corrected by applying a low-pass filter on homogeneous areas.

\vspace{5pt}
\noindent\textbf{Complex textures.} In addition, the model occasionally struggles with fine-grained details such as the text present in the upper right of \autoref{fig:limitation_car}. This limitation is likely due to limited text in the training data, as well as the model's reliance on blur maps and image propagation components, which might not effectively handle all complex textures.

\vspace{5pt}
\noindent\textbf{Computational complexity.} 
DaBiT is a high-performance, but relatively large, model with competitive inference speed. This allows for use in time-sensitive scenarios; however, the large size imposes high memory requirements and restricts deployment of the model on low-end hardware. Future work will aim to combat these drawbacks with model compression techniques such as model pruning optimizers like OBProx-SG~\cite{chen2021orthant} and knowledge distillation~\cite{hinton2015distilling,morris2023st}. 

\section{Conclusion}
\label{sec:conclusion}

This paper presents a novel approach to the relatively understudied problem of video focal deblurring. Our method utilizes a novel transformer and flow refocusing module, along with bidirectional image propagation, to remove focal blur from a given video sequence. We overcome the limitations of computational resources and speed by working with low-resolution inputs and embedding a super-resolution module in the pipeline. Additionally, we propose a new method to generate synthetic focal blur data, enabling the model to learn from a broader range of content. Compared to state-of-the-art video deblurring and super-resolution models on the proposed DAVIS-Blur test set, our method achieves superior results, particularly in detail preservation and enhancing temporal consistency. The model is also highly adaptable to real-world low-light footage, a common source of focusing issues. Future work will focus on generalizing this approach to other forms of blur, increasing the super-resolution scaling, and investigating if similar performance can be achieved at increased speed by a smaller model using model compression and knowledge distillation techniques.




{\small
\bibliographystyle{ieee_fullname}
\bibliography{egbib}

\begin{thebibliography}{10}\itemsep=-1pt

\bibitem{abuolaim2020defocus}
Abdullah Abuolaim and Michael~S Brown.
\newblock Defocus deblurring using dual-pixel data.
\newblock In {\em Computer Vision--ECCV 2020: 16th European Conference, Glasgow, UK, August 23--28, 2020, Proceedings, Part X 16}, pages 111--126. Springer, 2020.

\bibitem{anant2022artificial}
Nakarin Anantrasirichai and David Bull.
\newblock Artificial intelligence in the creative industries: a review.
\newblock {\em Artificial Intelligence Review}, 55(1):589--656, 2022.

\bibitem{bhat2023zoedepth}
Shariq~Farooq Bhat, Reiner Birkl, Diana Wofk, Peter Wonka, and Matthias M{\"u}ller.
\newblock Zoedepth: Zero-shot transfer by combining relative and metric depth.
\newblock {\em arXiv preprint arXiv:2302.12288}, 2023.

\bibitem{chan2021basicvsr}
Kelvin~CK Chan, Xintao Wang, Ke Yu, Chao Dong, and Chen~Change Loy.
\newblock Basicvsr: The search for essential components in video super-resolution and beyond.
\newblock In {\em Proceedings of the IEEE/CVF conference on computer vision and pattern recognition}, pages 4947--4956, 2021.

\bibitem{chan2022basicvsr++}
Kelvin~CK Chan, Shangchen Zhou, Xiangyu Xu, and Chen~Change Loy.
\newblock Basicvsr++: Improving video super-resolution with enhanced propagation and alignment.
\newblock In {\em Proceedings of the IEEE/CVF conference on computer vision and pattern recognition}, pages 5972--5981, 2022.

\bibitem{chen2022simple}
Liangyu Chen, Xiaojie Chu, Xiangyu Zhang, and Jian Sun.
\newblock Simple baselines for image restoration.
\newblock {\em arXiv preprint arXiv:2204.04676}, 2022.

\bibitem{chen2021orthant}
Tianyi Chen, Tianyu Ding, Bo Ji, Guanyi Wang, Yixin Shi, Jing Tian, Sheng Yi, Xiao Tu, and Zhihui Zhu.
\newblock Orthant based proximal stochastic gradient method for $l_1$-regularized optimization.
\newblock In {\em Machine Learning and Knowledge Discovery in Databases: European Conference, ECML PKDD 2020, Ghent, Belgium, September 14--18, 2020, Proceedings, Part III}, pages 57--73. Springer, 2021.

\bibitem{chen2023better}
Xuhai Chen, Jiangning Zhang, Chao Xu, Yabiao Wang, Chengjie Wang, and Yong Liu.
\newblock Better" cmos" produces clearer images: Learning space-variant blur estimation for blind image super-resolution.
\newblock In {\em Proceedings of the IEEE/CVF Conference on Computer Vision and Pattern Recognition}, pages 1651--1661, 2023.

\bibitem{chu2020learning}
Mengyu Chu, You Xie, Jonas Mayer, Laura Leal-Taix{\'e}, and Nils Thuerey.
\newblock Learning temporal coherence via self-supervision for gan-based video generation.
\newblock {\em ACM Transactions on Graphics (TOG)}, 39(4):75--1, 2020.

\bibitem{cui2023image}
Yuning Cui, Wenqi Ren, Xiaochun Cao, and Alois Knoll.
\newblock Image restoration via frequency selection.
\newblock {\em IEEE Transactions on Pattern Analysis and Machine Intelligence}, 2023.

\bibitem{dai2017deformable}
Jifeng Dai, Haozhi Qi, Yuwen Xiong, Yi Li, Guodong Zhang, Han Hu, and Yichen Wei.
\newblock Deformable convolutional networks.
\newblock In {\em Proceedings of the IEEE international conference on computer vision}, pages 764--773, 2017.

\bibitem{danier2022st}
Duolikun Danier, Fan Zhang, and David Bull.
\newblock {ST-MFNet}: A spatio-temporal multi-flow network for frame interpolation.
\newblock In {\em Proceedings of the IEEE/CVF Conference on Computer Vision and Pattern Recognition}, pages 3521--3531, 2022.

\bibitem{dosovitskiy2020image}
Alexey Dosovitskiy, Lucas Beyer, Alexander Kolesnikov, Dirk Weissenborn, Xiaohua Zhai, Thomas Unterthiner, Mostafa Dehghani, Matthias Minderer, Georg Heigold, Sylvain Gelly, et~al.
\newblock An image is worth 16x16 words: Transformers for image recognition at scale.
\newblock {\em arXiv preprint arXiv:2010.11929}, 2020.

\bibitem{dudhane2021burst}
Akshay Dudhane, Syed~Waqas Zamir, Salman Khan, Fahad~Shahbaz Khan, and Ming-Hsuan Yang.
\newblock Burst image restoration and enhancement.
\newblock In {\em CVPR}, 2022.

\bibitem{gao2020flow}
Chen Gao, Ayush Saraf, Jia-Bin Huang, and Johannes Kopf.
\newblock Flow-edge guided video completion.
\newblock In {\em Computer Vision--ECCV 2020: 16th European Conference, Glasgow, UK, August 23--28, 2020, Proceedings, Part XII 16}, pages 713--729. Springer, 2020.

\bibitem{hinton2015distilling}
Geoffrey Hinton, Oriol Vinyals, and Jeff Dean.
\newblock Distilling the knowledge in a neural network.
\newblock {\em arXiv preprint arXiv:1503.02531}, 2015.

\bibitem{huang2015bidirectional}
Yan Huang, Wei Wang, and Liang Wang.
\newblock Bidirectional recurrent convolutional networks for multi-frame super-resolution.
\newblock {\em Advances in neural information processing systems}, 28, 2015.

\bibitem{hyun2015generalized}
Tae Hyun~Kim and Kyoung Mu~Lee.
\newblock Generalized video deblurring for dynamic scenes.
\newblock In {\em Proceedings of the IEEE Conference on Computer Vision and Pattern Recognition}, pages 5426--5434, 2015.

\bibitem{jin2018learning}
Meiguang Jin, Givi Meishvili, and Paolo Favaro.
\newblock Learning to extract a video sequence from a single motion-blurred image.
\newblock In {\em Proceedings of the IEEE Conference on Computer Vision and Pattern Recognition}, pages 6334--6342, 2018.

\bibitem{jo2018deep}
Younghyun Jo, Seoung~Wug Oh, Jaeyeon Kang, and Seon~Joo Kim.
\newblock Deep video super-resolution network using dynamic upsampling filters without explicit motion compensation.
\newblock In {\em Proceedings of the IEEE conference on computer vision and pattern recognition}, pages 3224--3232, 2018.

\bibitem{kim2017dynamic}
Tae~Hyun Kim, Seungjun Nah, and Kyoung~Mu Lee.
\newblock Dynamic video deblurring using a locally adaptive blur model.
\newblock {\em IEEE transactions on pattern analysis and machine intelligence}, 40(10):2374--2387, 2017.

\bibitem{kingma2014adam}
Diederik~P Kingma and Jimmy Ba.
\newblock Adam: A method for stochastic optimization.
\newblock {\em arXiv preprint arXiv:1412.6980}, 2014.

\bibitem{kong2023efficient}
Lingshun Kong, Jiangxin Dong, Jianjun Ge, Mingqiang Li, and Jinshan Pan.
\newblock Efficient frequency domain-based transformers for high-quality image deblurring.
\newblock In {\em Proceedings of the IEEE/CVF Conference on Computer Vision and Pattern Recognition}, pages 5886--5895, 2023.

\bibitem{kupyn2019deblurgan}
Orest Kupyn, Tetiana Martyniuk, Junru Wu, and Zhangyang Wang.
\newblock Deblurgan-v2: Deblurring (orders-of-magnitude) faster and better.
\newblock In {\em Proceedings of the IEEE/CVF international conference on computer vision}, pages 8878--8887, 2019.

\bibitem{li2023simple}
Dasong Li, Xiaoyu Shi, Yi Zhang, Ka~Chun Cheung, Simon See, Xiaogang Wang, Hongwei Qin, and Hongsheng Li.
\newblock A simple baseline for video restoration with grouped spatial-temporal shift.
\newblock In {\em Proceedings of the IEEE/CVF Conference on Computer Vision and Pattern Recognition}, pages 9822--9832, 2023.

\bibitem{li2020mucan}
Wenbo Li, Xin Tao, Taian Guo, Lu Qi, Jiangbo Lu, and Jiaya Jia.
\newblock Mucan: Multi-correspondence aggregation network for video super-resolution.
\newblock In {\em Computer Vision--ECCV 2020: 16th European Conference, Glasgow, UK, August 23--28, 2020, Proceedings, Part X 16}, pages 335--351. Springer, 2020.

\bibitem{li2022towards}
Zhen Li, Cheng-Ze Lu, Jianhua Qin, Chun-Le Guo, and Ming-Ming Cheng.
\newblock Towards an end-to-end framework for flow-guided video inpainting.
\newblock In {\em Proceedings of the IEEE/CVF conference on computer vision and pattern recognition}, pages 17562--17571, 2022.

\bibitem{liang2024vrt}
Jingyun Liang, Jiezhang Cao, Yuchen Fan, Kai Zhang, Rakesh Ranjan, Yawei Li, Radu Timofte, and Luc Van~Gool.
\newblock Vrt: A video restoration transformer.
\newblock {\em IEEE Transactions on Image Processing}, 2024.

\bibitem{liang2022rvrt}
Jingyun Liang, Yuchen Fan, Xiaoyu Xiang, Rakesh Ranjan, Eddy Ilg, Simon Green, Jiezhang Cao, Kai Zhang, Radu Timofte, and Luc~V Gool.
\newblock Recurrent video restoration transformer with guided deformable attention.
\newblock {\em Advances in Neural Information Processing Systems}, 35:378--393, 2022.

\bibitem{lin2024bvi}
Rui Lin, Nakarin Anantrasirichai, Guanghui Huang, Jiacheng Lin, Qiyuan Sun, Alena Malyugina, and David~R Bull.
\newblock {BVI-RLV}: A fully registered dataset and benchmarks for low-light video enhancement.
\newblock {\em arXiv preprint arXiv:2407.03535}, 2024.

\bibitem{liu2022learning}
Chengxu Liu, Huan Yang, Jianlong Fu, and Xueming Qian.
\newblock Learning trajectory-aware transformer for video super-resolution.
\newblock In {\em Proceedings of the IEEE/CVF conference on computer vision and pattern recognition}, pages 5687--5696, 2022.

\bibitem{liu2021fuseformer}
Rui Liu, Hanming Deng, Yangyi Huang, Xiaoyu Shi, Lewei Lu, Wenxiu Sun, Xiaogang Wang, Jifeng Dai, and Hongsheng Li.
\newblock Fuseformer: Fusing fine-grained information in transformers for video inpainting.
\newblock In {\em Proceedings of the IEEE/CVF international conference on computer vision}, pages 14040--14049, 2021.

\bibitem{ma2020bvi}
Di Ma, Fan Zhang, and David Bull.
\newblock {BVI-DVC}: A training database for deep video compression.
\newblock {\em IEEE Transactions on Multimedia}, pages 1--1, 2021.

\bibitem{ma2020mfrnet}
Di Ma, Fan Zhang, and David~R Bull.
\newblock {MFRNet}: a new {CNN} architecture for post-processing and in-loop filtering.
\newblock {\em IEEE Journal of Selected Topics in Signal Processing}, 15(2):378--387, 2020.

\bibitem{morris2023st}
Crispian Morris, Duolikun Danier, Fan Zhang, Nantheera Anantrasirichai, and David~R Bull.
\newblock {ST-MFNet mini}: Knowledge distillation-driven frame interpolation.
\newblock In {\em 2023 IEEE International Conference on Image Processing (ICIP)}, pages 1045--1049. IEEE, 2023.

\bibitem{perazzi2016benchmark}
Federico Perazzi, Jordi Pont-Tuset, Brian McWilliams, Luc Van~Gool, Markus Gross, and Alexander Sorkine-Hornung.
\newblock A benchmark dataset and evaluation methodology for video object segmentation.
\newblock In {\em Proceedings of the IEEE conference on computer vision and pattern recognition}, pages 724--732, 2016.

\bibitem{ranftl_towards_2020}
René Ranftl, Katrin Lasinger, David Hafner, Konrad Schindler, and Vladlen Koltun.
\newblock Towards {Robust} {Monocular} {Depth} {Estimation}: {Mixing} {Datasets} for {Zero}-shot {Cross}-dataset {Transfer}, Aug. 2020.
\newblock arXiv:1907.01341 [cs].

\bibitem{ruan2022learning}
Lingyan Ruan, Bin Chen, Jizhou Li, and Miuling Lam.
\newblock Learning to deblur using light field generated and real defocus images.
\newblock In {\em Proceedings of the IEEE/CVF Conference on Computer Vision and Pattern Recognition}, pages 16304--16313, 2022.

\bibitem{shang2023joint}
Wei Shang, Dongwei Ren, Yi Yang, Hongzhi Zhang, Kede Ma, and Wangmeng Zuo.
\newblock Joint video multi-frame interpolation and deblurring under unknown exposure time.
\newblock In {\em Proceedings of the IEEE/CVF Conference on Computer Vision and Pattern Recognition}, pages 13935--13944, 2023.

\bibitem{son2021single}
Hyeongseok Son, Junyong Lee, Sunghyun Cho, and Seungyong Lee.
\newblock Single image defocus deblurring using kernel-sharing parallel atrous convolutions.
\newblock In {\em Proceedings of the IEEE/CVF International Conference on Computer Vision}, pages 2642--2650, 2021.

\bibitem{teed2020raft}
Zachary Teed and Jia Deng.
\newblock Raft: Recurrent all-pairs field transforms for optical flow.
\newblock In {\em Computer Vision--ECCV 2020: 16th European Conference, Glasgow, UK, August 23--28, 2020, Proceedings, Part II 16}, pages 402--419. Springer, 2020.

\bibitem{tian2020tdan}
Yapeng Tian, Yulun Zhang, Yun Fu, and Chenliang Xu.
\newblock Tdan: Temporally-deformable alignment network for video super-resolution.
\newblock In {\em Proceedings of the IEEE/CVF conference on computer vision and pattern recognition}, pages 3360--3369, 2020.

\bibitem{tsai2022stripformer}
Fu-Jen Tsai, Yan-Tsung Peng, Yen-Yu Lin, Chung-Chi Tsai, and Chia-Wen Lin.
\newblock Stripformer: Strip transformer for fast image deblurring.
\newblock In {\em European Conference on Computer Vision}, pages 146--162. Springer, 2022.

\bibitem{wang2004image}
Zhou Wang, Alan~C Bovik, Hamid~R Sheikh, and Eero~P Simoncelli.
\newblock Image quality assessment: from error visibility to structural similarity.
\newblock {\em IEEE transactions on image processing}, 13(4):600--612, 2004.

\bibitem{xu2018youtube}
Ning Xu, Linjie Yang, Yuchen Fan, Jianchao Yang, Dingcheng Yue, Yuchen Liang, Brian Price, Scott Cohen, and Thomas Huang.
\newblock Youtube-vos: Sequence-to-sequence video object segmentation.
\newblock In {\em Proceedings of the European conference on computer vision (ECCV)}, pages 585--601, 2018.

\bibitem{xu2019deep}
Rui Xu, Xiaoxiao Li, Bolei Zhou, and Chen~Change Loy.
\newblock Deep flow-guided video inpainting.
\newblock In {\em Proceedings of the IEEE/CVF Conference on Computer Vision and Pattern Recognition}, pages 3723--3732, 2019.

\bibitem{xu2024videogigagan}
Yiran Xu, Taesung Park, Richard Zhang, Yang Zhou, Eli Shechtman, Feng Liu, Jia-Bin Huang, and Difan Liu.
\newblock Videogigagan: Towards detail-rich video super-resolution.
\newblock {\em arXiv preprint arXiv:2404.12388}, 2024.

\bibitem{depthanything}
Lihe Yang, Bingyi Kang, Zilong Huang, Xiaogang Xu, Jiashi Feng, and Hengshuang Zhao.
\newblock Depth anything: Unleashing the power of large-scale unlabeled data.
\newblock {\em arXiv:2401.10891}, 2024.

\bibitem{youk2024fma}
Geunhyuk Youk, Jihyong Oh, and Munchurl Kim.
\newblock Fma-net: Flow-guided dynamic filtering and iterative feature refinement with multi-attention for joint video super-resolution and deblurring.
\newblock {\em arXiv preprint arXiv:2401.03707}, 2024.

\bibitem{zamir2022restormer}
Syed~Waqas Zamir, Aditya Arora, Salman Khan, Munawar Hayat, Fahad~Shahbaz Khan, and Ming-Hsuan Yang.
\newblock Restormer: Efficient transformer for high-resolution image restoration.
\newblock In {\em Proceedings of the IEEE/CVF conference on computer vision and pattern recognition}, pages 5728--5739, 2022.

\bibitem{zamir2021multi}
Syed~Waqas Zamir, Aditya Arora, Salman Khan, Munawar Hayat, Fahad~Shahbaz Khan, Ming-Hsuan Yang, and Ling Shao.
\newblock Multi-stage progressive image restoration.
\newblock In {\em Proceedings of the IEEE/CVF conference on computer vision and pattern recognition}, pages 14821--14831, 2021.

\bibitem{zhang2021video}
Fan Zhang, Di Ma, Chen Feng, and David~R Bull.
\newblock Video compression with {CNN-based} postprocessing.
\newblock {\em IEEE MultiMedia}, 28(4):74--83, 2021.

\bibitem{zhang2022flow}
Kaidong Zhang, Jingjing Fu, and Dong Liu.
\newblock Flow-guided transformer for video inpainting.
\newblock In {\em European Conference on Computer Vision}, pages 74--90. Springer, 2022.

\bibitem{zhao2022united}
Wenda Zhao, Fei Wei, You He, and Huchuan Lu.
\newblock United defocus blur detection and deblurring via adversarial promoting learning.
\newblock In {\em European Conference on Computer Vision}, pages 569--586. Springer, 2022.

\bibitem{zhong2023blur}
Zhihang Zhong, Mingdeng Cao, Xiang Ji, Yinqiang Zheng, and Imari Sato.
\newblock Blur interpolation transformer for real-world motion from blur.
\newblock In {\em Proceedings of the IEEE/CVF Conference on Computer Vision and Pattern Recognition}, pages 5713--5723, 2023.

\bibitem{zhou2023propainter}
Shangchen Zhou, Chongyi Li, Kelvin~CK Chan, and Chen~Change Loy.
\newblock Propainter: Improving propagation and transformer for video inpainting.
\newblock In {\em Proceedings of the IEEE/CVF International Conference on Computer Vision}, pages 10477--10486, 2023.

\bibitem{zhou2023upscale}
Shangchen Zhou, Peiqing Yang, Jianyi Wang, Yihang Luo, and Chen~Change Loy.
\newblock Upscale-a-video: Temporal-consistent diffusion model for real-world video super-resolution.
\newblock {\em arXiv preprint arXiv:2312.06640}, 2023.

\bibitem{zhu2019deformable}
Xizhou Zhu, Han Hu, Stephen Lin, and Jifeng Dai.
\newblock Deformable convnets v2: More deformable, better results.
\newblock In {\em Proceedings of the IEEE/CVF conference on computer vision and pattern recognition}, pages 9308--9316, 2019.

\end{thebibliography}
}




\end{document}